\documentclass[letterpaper, 10 pt, conference]{ieeeconf}  

\IEEEoverridecommandlockouts                              

\usepackage{graphicx}
\usepackage{cite}
\usepackage{color}
\usepackage{amsmath,amssymb} 
\usepackage{epsfig} 
\usepackage{multirow}
\usepackage{hyperref}
\usepackage{bm}
\usepackage{caption}
\usepackage{copyrightbox}
\usepackage{soul,xcolor} 

\title{\LARGE \bf
Maneuver-based Anchor Trajectory Hypotheses at Roundabouts
}

\author{Mohamed Hasan$^{1}$, Evangelos Paschalidis$^{2}$, Albert Solernou$^{2}$, He Wang$^{1}$,\\ Gustav Markkula$^{2}$  and Richard Romano$^{2}$
\thanks{$^{1}$ School of Computing, University of Leeds}%
\thanks{$^{2}$ Institute for Transport Studies, University of Leeds}%
}

\newcommand{\quotes}[1]{``#1''}

\begin{document}

\setstcolor{blue}

\maketitle
\thispagestyle{empty}
\pagestyle{empty}

\begin{abstract}

Predicting future behavior of the surrounding vehicles is crucial for self-driving platforms to safely navigate through other traffic. This is critical when making decisions like crossing an unsignalized intersection. We address the problem of vehicle motion prediction in a challenging roundabout environment by learning from human driver data. We extend existing recurrent encoder-decoder models to be advantageously combined with anchor trajectories to predict vehicle behaviors on a roundabout. Drivers' intentions are encoded by a set of maneuvers that correspond to semantic driving concepts. Accordingly, our model employs a set of maneuver-specific anchor trajectories that cover the space of possible outcomes at the roundabout. The proposed model can output a multi-modal distribution over the predicted future trajectories based on the maneuver-specific anchors.  We evaluate our model using the public RounD dataset and the experiment results show the effectiveness of the proposed maneuver-based anchor regression in improving prediction accuracy, reducing the average RMSE to 28\% less than the best baseline. Our code is available at https://github.com/m-hasan-n/roundabout.

\end{abstract}

\section{INTRODUCTION}

Predicting future agent states is a crucial task for robot planning and control in general \cite{chai2019multipath}. In this paper, we address this problem for self-driving vehicles to be able to predict behaviors of surrounding vehicles, which is vital for safe and efficient driving. For example, it is important when deciding whether to yield to a coming vehicle or to merge into traffic. 

To safely drive vehicles and navigate through other traffic, human drivers need to anticipate the intentions of other road users. Experienced human drivers are able to confidently infer future behaviors of the surrounding vehicles \cite{zyner2019naturalistic}. This is critical when making decisions like crossing an unsignalized intersection. 

In order to build fully autonomous vehicles, they should be able to predict the future states of its environment and respond appropriately. Predicting the behavior of human drivers is necessary for such autonomous platforms to share the road with humans \cite{gupta2018social}. More importantly, being able to predict human driver behaviors will lead to \textit{human-like} responding strategies, which is crucial for human road users to predict the autonomous vehicle behavior. While human drivers can do this prediction seamlessly, it is still an open problem for self-driving cars.

Vehicle motion prediction is challenging due to the inherent uncertainty about latent variables such as the motivations of road users. Additionally, driver behavior tends to be multi-modal: a driver could make one of many decisions under the same traffic circumstances \cite{deo2018convolutional}. 

Although this problem has been extensively studied on highways or on highly structured intersections, fewer works addressed un-signalized intersections like roundabouts \cite{zyner2018recurrent}. Roundabouts are popular in urban areas as they do not require expensive traffic lights, improve safety and have higher throughput than a four-way stop.  Being able to safely navigate these highly dynamic scenarios is critical for autonomous vehicles to function properly. 

We focus on addressing the problem of trajectory prediction of the human-driven vehicles surrounding an autonomous vehicle in a roundabout environment. In line with the existing work \cite{alahi2016social, deo2018convolutional}, this is achieved by iteratively predicting the trajectory of each surrounding agent around the autonomous one. We call the subject surrounding vehicle to be predicted as \quotes{\textit{ego}} and the vehicles around the ego vehicle as \quotes{\textit{neighbors}}. One concrete task to this end is: given the observed motion trajectories of natural human drivers (position and heading of the past e.g. \(2 \,s\)) of an ego vehicle and all its neighbor vehicles on a roundabout, to predict the future trajectory of this ego vehicle. We seek a predictive model that outputs a multi-modal distribution over the predicted trajectory.

We propose a model that employs a fixed set of anchor trajectories that cover the space of possible outcomes at the roundabout environment. The anchor trajectories are associated with a corresponding set of maneuvers that model the (short-term) intention of human drivers. These maneuvers (intentions) represent semantic concepts like \quotes{slow down} and \quotes{advance to a specific zone of the roundabout}.  

Our model can hierarchically factor the uncertainty inherent in the prediction process. First, the maneuver prediction captures the uncertainty about the driver's intention and is encoded as a distribution over a set of discrete intentions. We model a discrete set of anchor trajectories corresponding to the maneuver set covering the space of modelled human intentions. Second, given the intended maneuvers and the corresponding anchor trajectory, the predicted trajectory models the uncertainty performing such maneuvers by estimating a maneuver-specific residual from the given anchor trajectory. 

Thus, we introduce an encoder-decoder based model for vehicle motion prediction that can evaluate the likelihood of each maneuver and regress a maneuver-specific future trajectory segment as a residual from the corresponding anchor trajectory, providing a multi-modal distribution over the predicted future trajectories based on the maneuver-specific anchors.

The contributions of our work are three fold: (1) incorporating anchor trajectories into recurrent models to predict vehicle behaviors, (2) parameterization of the whole maneuver space in a roundabout by maneuver-specific anchor trajectories, and (3) extending a pooling strategy to account for vehicle poses that are predicted as a multi-variate Gaussian distribution.


\section{Related Work}
\textbf{Recurrent networks for trajectory prediction.}
Recurrent Neural Networks (RNNs) represent a rich class of dynamic models which extend feedforward networks for sequence generation \cite{graves2013generating} in diverse domains like speech recognition \cite{chorowski2014end}, machine translation \cite{chung2015recurrent} and image captioning \cite{vinyals2015show}. Motion prediction is considered as a sequence generation task. Hence, a number of RNN-based approaches have been proposed for trajectory prediction \cite{alahi2016social, gupta2018social, deo2018convolutional, zyner2019naturalistic, altche2017lstm, park2018sequence, xue2018ss}  of pedestrians and vehicles. In their seminal work Social LSTM (Long-Short Term Memory), Alahi et al. \cite{alahi2016social} extended RNNs to human trajectory prediction using a social pooling layer that models nearby pedestrians. Gupta et al. proposed a Generative Adversarial Networks \cite{goodfellow2014generative} (GAN): a RNN Encoder-Decoder generator and a RNN based encoder discriminator, to predict socially-acceptable multimodal pedestrian trajectories \cite{gupta2018social}. Social LSTM approach was further improved in \cite{deo2018convolutional} by using convolutional social pooling applied to vehicle motion prediction on highways. LSTM network is also used to predict the location of vehicles in an occupancy grid \cite{kim2017probabilistic} at different future intervals. Convolutional networks and LSTM were combined to predict multi-modal trajectories for an agent on a bird's eye view image \cite{bhattacharyya2018accurate}. The majority of the vehicle prediction literature addressed the problem on highways and structured intersections, whereas we focus on the vehicle trajectory prediction on roundabouts.  

\textbf{Multimodal predictive distribution.}
Deo et al. proposed a model \cite{deo2018convolutional} that outputs a multi-modal predictive distribution over future trajectories based on maneuver classes. Three lateral and two longitudinal maneuver classes have been considered. MultiPath model \cite{chai2019multipath} can predict a discrete distribution over a set of future state-sequence anchors and output multi-modal future distributions. The GAN based encoder-decoder architecture in \cite{gupta2018social} encourages diverse multimodal predictions of pedestrian trajectories with the introduced variety loss. A multi-modal probabilistic prediction approach was presented in \cite{hu2019multi} based on a Conditional
Variational Autoencoder and is capable of jointly predicting sequential motions of each pair of interacting vehicles. Zyner et al. \cite{zyner2019naturalistic} presented a model based on RNN with a mixture density network output layer, for predicting driver intent at urban single-lane roundabouts through multi-modal trajectory prediction with uncertainty. Neither maneuver recognition nor anchor-based regression were considered in \cite{hu2019multi} or \cite{zyner2019naturalistic} where the roundabout problem was addressed. 

\textbf{Anchor trajectories.} 
The concept of predefined anchors has been effectively applied in machine learning and computer vision to handle multi-modal problems \cite{bishop2006pattern, erhan2014scalable, yang2012articulated}. These approaches predict the likelihood of anchors as well as continuous refinements of state conditioned on these anchors. For the sake of vehicle trajectory prediction, MultiPath model \cite{chai2019multipath} employs a fixed set of trajectory anchors that are found in the training data via unsupervised learning. At inference time, the model predicts a discrete distribution over the anchors and, for each anchor, regresses offsets from anchor waypoints along with uncertainties. Considering the set of anchors as the driver intents, these were not associated with semantic concepts like  \quotes{slow down}  or \quotes{lane change}. Phan-Minh et. al introduced CoverNet \cite{phan2020covernet} for multimodal, probabilistic trajectory prediction for urban driving.  The trajectory prediction problem was framed as a classification over a diverse set of trajectories \cite{boulton2020motion}. The trajectory set was structured to cover the state space, and eliminated physically impossible trajectories. 

Our work contrasts with the previous approaches in the following aspects: (1) we design the anchor trajectories to correspond to the human driver maneuvers, (2) our model is based on recurrent (not convolutional like \cite{chai2019multipath}, \cite{phan2020covernet} and  \cite{boulton2020motion}) networks to encode the trajectory sequence history and decode its future prediction, and (3) it can capture the inter dependencies between all the participating vehicles (not being restricted by a specific grid size like \cite{alahi2016social} and \cite{deo2018convolutional}). 

\section{RounD Dataset}

The RounD dataset \cite{rounDdataset} is a new dataset of naturalistic road user trajectories recorded at German roundabouts. Traffic was recorded using a drone at three different locations, and the trajectory for each road user and its type was extracted. Using state-of-the-art computer vision algorithms, the positional error is typically less than 10 centimetres. The dataset provides \(X\) and \(Y\) position, velocity and acceleration, heading, lateral and longitudinal velocity and accelerations of the tracked objects in a local coordinate system, recorded at 25 Hz. In addition, the RounD dataset contains the dimensions (length and width) and class (e.g. car, truck, pedestrian) of the tracked objects. 

\begin{figure}
\vspace{2mm}
\centering 
\includegraphics[scale=0.245]{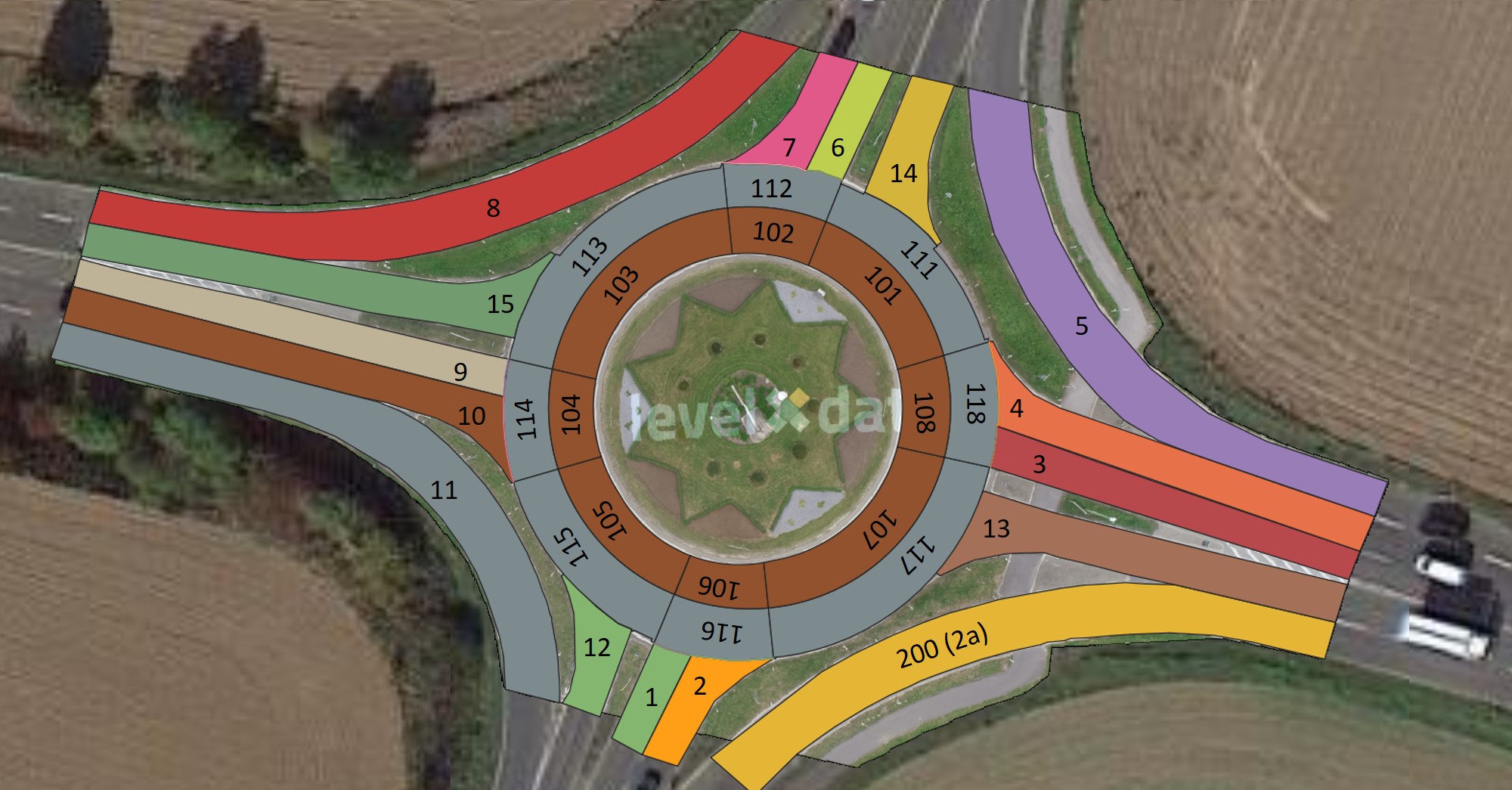}
\caption{The roundabout zones after segmentation.} 
\label{fig:lanes}
\vspace{-5mm}
\end{figure}

\subsection{Dataset Processing}

The roundabout environment was segmented -following \cite{Paschalidis2021estimation}- into zones that include: the entrance lanes, the main roundabout areas, the conflict zones and the exit lanes (Fig. \ref{fig:lanes}). To focus on the vehicle trajectory prediction at the roundabout scenario, we refined the dataset to exclude trajectories\st{:} of pedestrian, bicycle and motorcycle classes, and in none-relevant lanes (IDs 5,8,11 and 200 in Fig. \ref{fig:lanes}). The vehicle behavior long before it enters the roundabout or long after it leaves the roundabout is out of the scope of the paper. Therefore, we additionally removed the trajectory segments that are farther away from the roundabout (entries/exits) than the vehicle length.

\section{Problem Formulation}

Inline with the previous work \cite{deo2018convolutional, chai2019multipath}, We formulate the problem of vehicle trajectory prediction by estimating the probability distribution of the possible future trajectories of an ego vehicle conditioned on its track history and on the track histories of the vehicles around it, at each discrete time step \(t\). Thus, given observations \(X\) of past trajectories of the ego and its neighbor vehicles in a scene, our model seeks to provide a distribution over future trajectories of the ego vehicle \(P(Y|X)\).

\subsection{Inputs and Outputs} 
The inputs to our model are track histories:
\begin{align}
\begin{split}
X = [x^{(t-t_h)}, ..., x^{(t-1)}, x^{(t)}]
\end{split}
\end{align}
where \(t_h\) is a fixed (history) time horizon, and at any time instant \(t\),
\begin{align}
\begin{split}
x^{(t)} = [x_0^{(t)}, y_0^{(t)}, \theta_0^{(t)}, x_1^{(t)}, y_1^{(t)}, \theta_1^{(t)}, ..., x_n^{(t)}, y_n^{(t)}, \theta_n^{(t)}]
\end{split}
\end{align}
represents the pose of the ego vehicle (subscript \(0\)), and the surrounding vehicles (subscripts \(1\) to \(n\)). The pose is given by the position (\(x\) and \(y\) coordinates) and orientation \(\theta\). We assume that the sensors/computers on-board the autonomous vehicle can measure/compute the pose of the ego and neighbor vehicles. The output of the model is a probability distribution:
\begin{align}
\begin{split}
Y = [y^{(t+1)}, ..., y^{(t+t_f)}]
\end{split}
\end{align}
where \(t_f\) is a fixed (future) time horizon, and:
\begin{align}
\begin{split}
y^{(t)} = [x_0^{(t)}, y_0^{(t)}, \theta_0^{(t)}]
\end{split}
\end{align}
represents the future pose of the ego vehicle being predicted at any time instant $t$.

\subsection{Multi-modal Probabilistic Trajectory Prediction}

The multi-modal distribution over future trajectories can be hierarchically factorized \cite{chai2019multipath}. First, estimate the intent uncertainty over the set of maneuver classes, and get the corresponding anchor trajectory. For example, this may be the uncertainty about advancing to a specific zone of the roundabout with an intended speed profile (e.g. slowing down). Second, given the intended maneuvers and their corresponding anchor, predict the vehicle trajectory as a residual (offset) from the given anchor trajectory.  
 
Maneuvers are chosen to semantically represent how human drivers may plan their decisions. Drivers may first decide a target zone, then choose some acceleration to reach this target according to the traffic situation. Accordingly, intentions of drivers are modelled using two discrete sets of maneuver types: (1) location-wise maneuvers \(M_l=\{m_{l_p}\}_{p=1}^P\), and (2) acceleration-wise maneuvers \(M_a=\{m_{a_q}\}_{q=1}^Q\). We model the uncertainty over each discrete set of maneuvers using a softmax distribution. For example, the distribution over \(M_l\) is given by: 
\begin{align}
\begin{split}
P(m_{l_p}|X)=\frac{exp(f_{l_p}(X))}{\sum_i exp(f_{l_i}(X))} 
\end{split}
\end{align}
where \(f_{l_p}(X)\) is the output of a deep neural network whose input is the encoding of the history trajectories.

Correspondingly, we model a discrete set of \(K\) maneuver-specific anchor trajectories \(A=\{a_k\}_{k=1}^K\). Each anchor trajectory is modelled as a sequence of future poses of the ego vehicle: 
\begin{align}
\begin{split}
a_k = [y_k^{(t+1)}, ..., y_k^{(t+t_f)}] 
\end{split}
\label{eq:anch}
\end{align}
Anchor trajectories are defined to cover the joint maneuver space, i.e. \(K = PQ\). The distribution over the anchor trajectories can be computed from the probabilities of the maneuver classes as follows:
\begin{align}
\begin{split}
P(a_k|X) = P(m_{l_p}|X) P(m_{a_q}|X)
\end{split}
\label{eq:anch_probability}
\end{align}

Given the intended maneuvers and the corresponding anchor trajectory, the uncertainty over the future trajectory is modelled as a Gaussian distribution:
\begin{align}
\begin{split}
P_{\Theta_k}(Y|a_k, X) = N(Y|a_k+\mu_k(X), \Sigma_k(X))
\end{split}
\end{align}
where in the Gaussian distribution mean \(a_k+\mu_k(X)\), the term \(\mu_k(X)\) represents an offset from the given anchor trajectory \(a_k\).  This allows the model to refine the static maneuver-specific anchor trajectories to the current context \cite{chai2019multipath}, with variations coming from the history trajectories of the vehicles in the scene. This distribution is parameterized by \cite{deo2018convolutional}:  
\begin{align}
\begin{split}
\Theta_k = [\Theta_k^{(t+1)}, ..., \Theta_k^{(t+t_f)}]
\end{split}
\end{align}
which are the parameters of a multi-variate Gaussian distribution at each time step in the future. At any time \(t\), this is given by the Gaussian parameters: \( \Theta_k^t = [\mu_k^{(t)}, \Sigma_k^{(t)}]\), corresponding to the mean and variance of the future vehicle pose as an offset from the anchor trajectory pose at the same time step.

The multi-modal output conditional distribution over future trajectories can now be expanded in terms of the anchors as:
\begin{align}
\begin{split}
P(Y|X) = \sum_k P(a_k|X) P_{\Theta_k}(Y|a_k, X)
\end{split}
\end{align}
which yields to a Gaussian Mixture Model distribution (GMM) \cite{chai2019multipath}. The mixture weights are defined by the probabilities of the anchors.

\section{Maneuver-based Anchor Trajectories}
\subsection{Maneuver Classes}
\label{sec:mnvrs}
We consider two types of maneuvers: location-wise \(M_l\) and acceleration-wise \(M_a\) maneuvers. The first type defines the intended future location on the roundabout (\(P\) = 8 classes), whereas the second type defines the intended speed-profile (\(Q\) = 3 classes) when advancing to occupy this space.

\begin{figure}
\vspace{2mm}
\centering 
\includegraphics[width=0.4\textwidth]{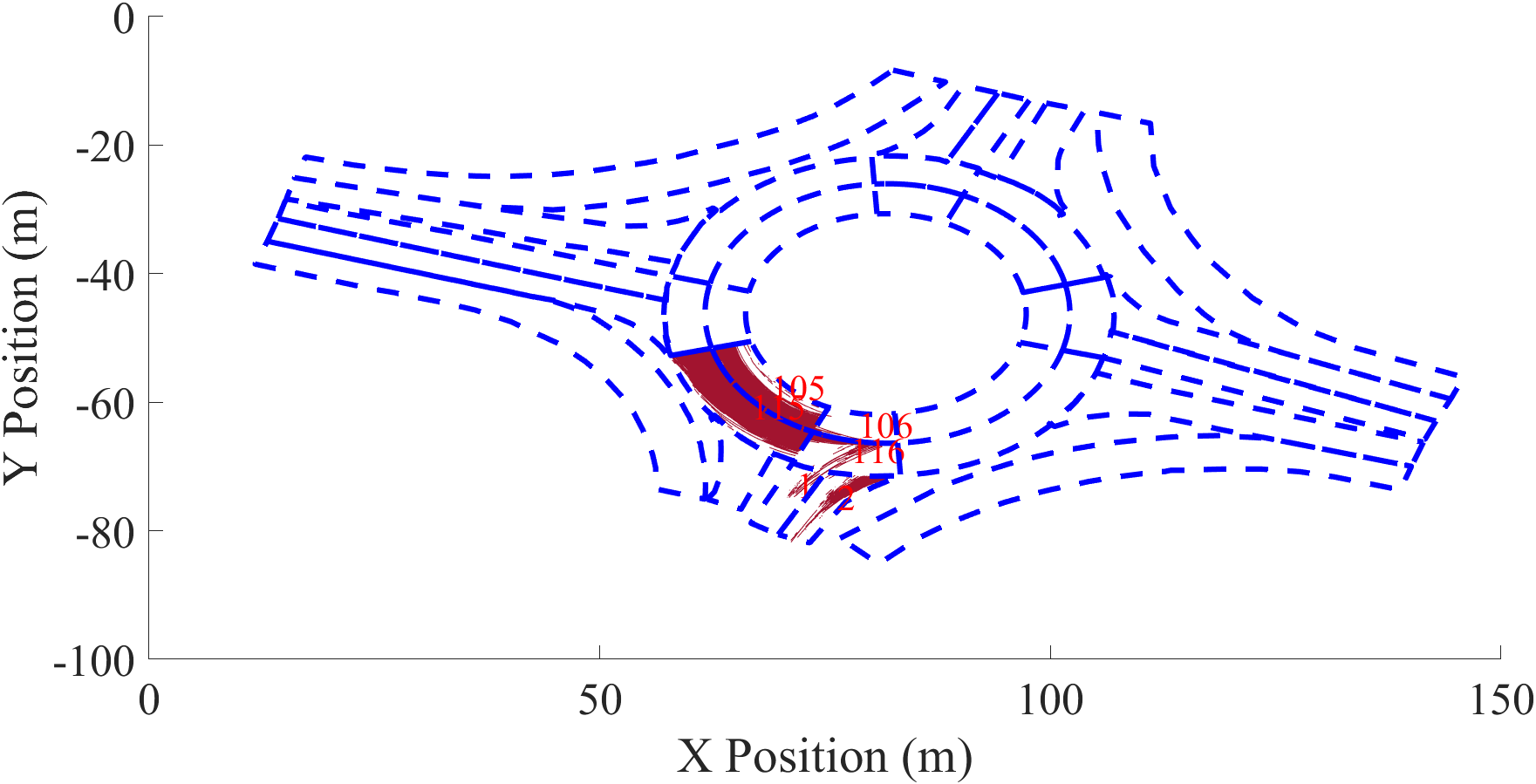}
\caption{An example of location-wise maneuver classes. The shown trajectories belong to same class vehicles, since their future will occupy the section grouping lanes 106 and 116.} 
\label{fig:intentions}
\vspace{-5mm}
\end{figure}

The circular infrastructure of the roundabout is segmented into 16 zones (101 to 108 and 111 to 118 as shown in Fig. \ref{fig:lanes}). We group these zones into eight sections which define the location-wise classes. For example, the trajectories shown in Fig. \ref{fig:intentions} are for vehicles belonging to the same class, as their future trajectories will occupy the section grouping the two zones 106 and 116. This includes vehicles: advancing from the previous section (the one grouping zones 105 and 115), entering the roundabout (from zones 1 and 2) and changing zones within the same section (between zones 106 and 116). Although this definition of location classes depends on the given roundabout segmentation, our model can cope with any given segmentation as long as it provides a reasonable number of well-distinguished classes. 

Three acceleration-wise maneuvers are considered: slowing down, constant speed and speeding up. The RounD dataset provides the acceleration of each vehicle. We annotated trajectory segments with an acceleration less/greater than \(-a\)/\(+a\) with the slowing/speeding class labels respectively while marking trajectory segments under the constant-speed class otherwise.  

According to the intended entry and exit lanes, a vehicle trajectory passes through a sequence of zones of the roundabout. This is accompanied by a sequence of acceleration intents (e.g. slowing down before the entrance or speeding up once accepting a gap). Accordingly, the pose of the vehicle at each time instant is annotated with the labels of the location and acceleration classes.       

\subsection{Anchor Trajectories}
Anchor trajectories are defined over the set of the joint location-acceleration maneuver space. This results in a total of \(K\) = 24 anchor trajectories corresponding to the grid of maneuvers, as for each of the eight spatial classes in Fig. \ref{fig:intentions}, we have three anchor trajectories correspond to this zone of the roundabout (one anchor for each acceleration class). 

Given a vehicle trajectory, the pose at each time \(t\) is annotated by two maneuver labels as given in Sec \ref{sec:mnvrs}. The future trajectory segment from \(t+1\) to \(t+t_f\) is extracted and annotated by the same maneuver labels. The vehicle poses of this segment are referenced to the pose at time \(t\). Repeating this process for all trajectories and for each time step, we end up with \(K\) trajectory sets, each of them includes a number of trajectory segments annotated by the same maneuver labels. 

\begin{figure}
\vspace{2mm}
\centering 
\includegraphics[width=0.26\textwidth]{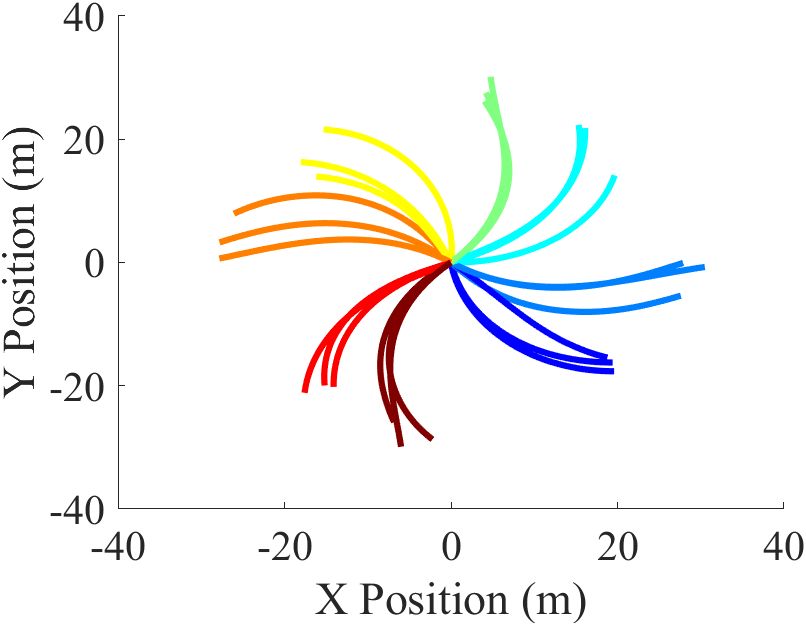}
\caption{The set of anchor trajectories. Each of the 8 location classes includes 3 anchors (corresponding to the acceleration classes) that are plotted with the same color. } 
\label{fig:anchors}
\vspace{-5mm}
\end{figure}

Anchor trajectories are obtained by finding the best representative trajectory of each set. This is done in our work by finding the mean trajectory of a subset of  empirically chosen candidates from each set. The resulting set of anchor trajectories employed in this work are shown in Fig. \ref{fig:anchors}. Each anchor trajectory is modelled as a sequence of future poses of the ego vehicle (as given by (\ref{eq:anch})) that starts with a zero pose and has a fixed number of \(t_f\) discrete time steps. 

\section{Model}

\begin{figure*}
\vspace{2mm}
\centering 
\includegraphics[width=0.8\textwidth]{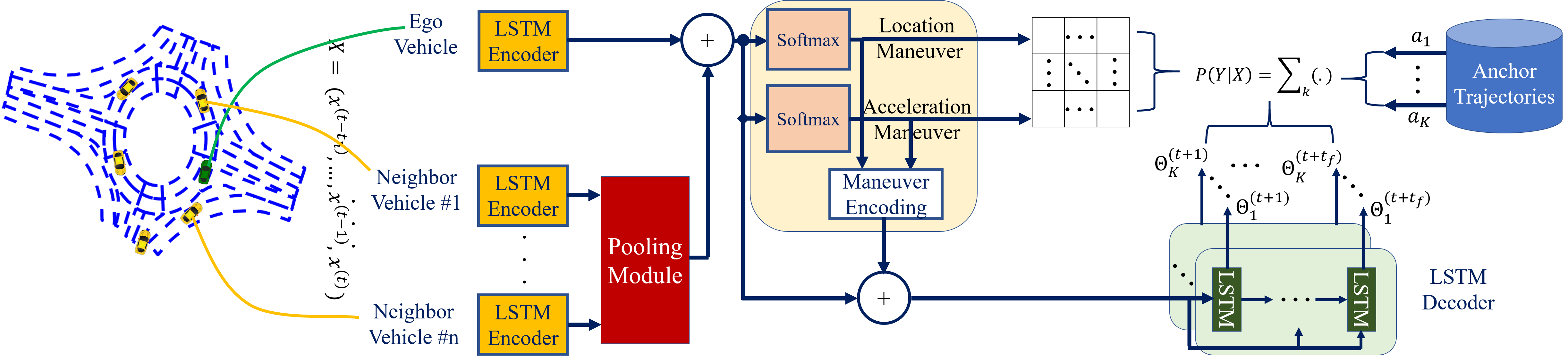}
\caption{Proposed model having the following modules: (1) encoder: learns the vehicle dynamics, (2) pooling: learns the interaction between vehicles, (3) maneuver: estimates maneuver class probabilities, and (4) decoder: outputs a multi-modal distribution of the ego vehicle future conditioned on the maneuver-based anchor trajectories.}
\label{fig:model}
\vspace{-4mm}
\end{figure*}

The proposed model is shown in Fig. \ref{fig:model} consisting of an LSTM encoder, a pooling module, a maneuver recognition module and an LSTM decoder.

\subsection{Encoder}
We encode the state of motion of each vehicle using an LSTM encoder. At any time \(t\), a sequence of \(t_h\) time steps of the track history is passed through the encoder. The LSTM states for each vehicle are updated frame by frame over the \(t_h\) past frames \cite{deo2018convolutional}. This is applied to all vehicles: the one whose future is being predicted (ego) and its surrounding vehicles (neighbors). The final LSTM (hidden) state of each vehicle is assumed to encode its motion dynamics, and the LSTM weights are shared across the sequences of all vehicles \cite{alahi2016social}.   

\subsection{Pooling Module}
\label{sec:pooling}
The interaction of vehicles in a given scene is captured by the pooling module. This is achieved by pooling the LSTM states of all the neighbor vehicles around the ego vehicle. The output is a pooling vector summarizing the context information. Social Pooling \cite{alahi2016social} and its convolutional extension \cite{deo2018convolutional} addressed this problem by proposing a grid based pooling scheme. However, this solution fails to capture global context, and may not suit the roundabout environment. Hence, we extend the pooling mechanism in \cite{gupta2018social} to model vehicle poses. 

First, at any time \(t\) the neighbor vehicles around the ego vehicle are identified. The relative pose between each neighbor and the ego vehicle is then computed, relative to the ego pose at \(t\). Second, the relative poses are concatenated with each vehicle’s LSTM hidden state, and processed independently by a Multi-Layer Perceptron (MLP). Finally, The MLP output is Max-pooled to compute the pooling vector of the ego vehicle. This method can capture the inter-dependencies among all vehicles, without being restricted to a specific grid size.

\subsection{Maneuver Recognition}

The maneuver recognition module consists of two softmax layers to recognize the location and acceleration classes. The input to this module is the LSTM state of the ego vehicle augmented by the pooling vector. Each softmax layer outputs the probability of each maneuver class. Probabilities output by the two layers are multiplied to calculate the probability for choosing each anchor trajectory (\(P(a_k|X)\) in (\ref{eq:anch_probability})). The maneuver encoding block in Fig. \ref{fig:model} encodes the output from each softmax layer into a one-hot vector. Both vectors are concatenated with the trajectory encoding and the resulting tensor is passed to the decoder module.

\subsection{Decoder}
An LSTM decoder is used to generate the conditional distribution over future trajectories of the ego vehicle. At any time \(t\), the decoder generates future trajectories over the next \(t_f\) steps in the form of multi-modal distribution. For each anchor trajectory, the decoder estimates an anchor-specific mean and variance residual of the vehicle pose.   

\section{Experiments and Results}

\subsection{Implementation Details}
We use the RounD dataset (consisting of 22 subsets recorded at three different locations) in our experiments. We use the data from the 20 subsets that were fully recorded at the same roundabout (Fig. \ref{fig:lanes}). The whole data are split into training (\(71\%\)), validation (\(10\%\)) and testing (\(19\%\)) sets. Each of the three sets is randomly sampled from all the employed subsets of RounD. Anchor trajectories are computed offline using the training set.

Vehicle trajectories are split into \(6 \,s\)-segments, where \(t_h=2 \,s\) track history and \(t_f=4 \,s\) prediction horizon are used. These \(6 \,s\)-segments are sampled at the dataset sampling rate of 25 Hz, and then downsampled by a factor of 4 before going to the LSTMs, to reduce the model complexity.

We implemented our proposed model based on the basic features of \cite{deo2018convolutional} using the same negative log likelihood loss. We extended these features to account for: (1) anchor-based prediction, (2) the pooling mechanism given in Sec. \ref{sec:pooling}, and (3) the multi-variate Gaussian distribution. The LSTM encoder and decoder have 32- and 64-dimensional states respectively. The fully connected layer used to embed the vehicle dynamics encoding has a size of 16. The MLP used in the pooling module has a size of 256 and is followed by batch-normalization and leaky-ReLU layers. The model is implemented using PyTorch \cite{paszke2017automatic} and trained end-to-end.

\subsection{Evaluation metric}

All results are reported in terms of the root of the mean squared error (RMSE) of the predicted trajectories with respect to the ground truth future trajectories, over the prediction horizon. Since the LSTM models generate multi-variate Gaussian distributions, the means of the Gaussian components are used for RMSE calculation.

\subsection{Baselines}
We characterised the performance of the following models in increasing complexity:
\begin{itemize}
  \item \textbf{2D}: vehicle trajectory is represented by the position (\(x\) and \(y\)), similar to \cite{alahi2016social, gupta2018social, deo2018convolutional}. Maneuvers and anchors are not included. 
  
  \item \textbf{3D}: vehicle trajectory is represented by the pose (position and heading \(\theta\)). Maneuvers and anchors are not included.  
  
  \item \textbf{3D-M}: adds the maneuver recognition module to the previous 3D model.
  
  \item \textbf{3D-A-P}: adds the anchors estimation to the previous model. This is our full proposed model. At the evaluation time, the model outputs the maximum \textit{a posteriori} probability (\emph{MAP}) trajectory estimate using the anchor trajectory having the maximum probability. 
  
  \item \textbf{3D-A-W}: a variation of our full proposed model. At evaluation time, the model outputs a \emph{weighted} sum of the MAP trajectory estimates from all anchors. The weights are defined by anchor probabilities \(P(a_k|X)\).
  
\end{itemize}

\subsection{Results}
RMSE results of the baseline models are reported in Table \ref{table:results}. Although the results of the 2D model are better than its counterpart 3D model at the short horizon (up to 1s), its prediction accuracy degrades with increasing prediction horizon. This shows how the heading of a vehicle is a valuable signal, especially when crossing a roundabout. Using the maneuver-recognition module in the 3D-M model improves the results further by reducing the RMSE average by nearly 24\(\%\).  

The full proposed model has been evaluated in two ways: (1) using the 3D-A-P model that predicts a single MAP trajectory estimate using the most-likely anchor trajectory, and (2) through the 3D-A-W model that outputs a GMM distribution, with mixture of weights defined by the probabilities of the anchors. Table \ref{table:results} shows that the latter improves the prediction accuracy over all other baselines. More importantly, comparing the results shows the effectiveness of generating a multi-modal distribution against a single best-estimate.    

\vspace{4mm}
\begin{center}
\begin{table}
\caption{Performance comparison of the proposed model in increasing complexity using the RMSE of the predicted trajectories (m) at different prediction horizons (s).} 
\centering 
\begin{tabular}{c c c c c c} 
\hline\hline 
Prediction\\ Horizon (s) & 2D & 3D & 3D-M & 3D-A-P & 3D-A-W\\ 
\hline \hline 
1 & 0.53  & 0.55  & 0.42 & 0.39 & \textbf{0.38}\\
2 & 1.62  & 1.32 & 0.87 & 0.84  & \textbf{0.80}\\
3 & 3.10 & 2.50 & 1.95 & 1.87 & \textbf{1.76}\\ 
4 & 4.92 & 4.02 & 3.31 & 3.27 & \textbf{3.08}\\
\hline\hline
\end{tabular}
\label{table:results}
\end{table}
\vspace{-7mm}
\end{center}

\vspace{-3mm}
\section{CONCLUSIONS}
We demonstrated how existing encoder-decoder based models can be advantageously combined with anchor trajectories to predict vehicle behaviors. We also argue for the benefit of using anchor trajectories that are based on human drivers' intentions. The effectiveness of the proposed model employing multi-modal estimation and anchor-based regression, has been proved through experiments on the public RounD dataset. One limitation of the proposed approach is its pure dependence on vehicle track histories. Contextual information like road semantics and visual cues can be used to improve the prediction accuracy. Incorporating such information into the current model will be addressed in our future work.

\addtolength{\textheight}{-12cm}   




\section*{ACKNOWLEDGMENT}
The work described in this paper is supported by VeriCAV project, which is funded by the Centre for Connected and Autonomous Vehicles, via Innovate UK (Grant number 104527). This paper is published with kind permission from the VeriCAV consortium: Horiba MIRA, Aimsun, University of Leeds, and Connected Places Catapult. This work was undertaken on ARC4, part of the High Performance Computing facilities at the University of Leeds, UK.


\bibliographystyle{IEEEtran}
\bibliography{IEEEabrv,my_ref.bib}

\end{document}